\documentclass[letterpaper]{article} 
\usepackage{aaai24}  
\usepackage{times}  
\usepackage{helvet}  
\usepackage{courier}  
\usepackage[hyphens]{url}  
\usepackage{graphicx} 
\usepackage{natbib}  
\usepackage{caption} 
\usepackage{algorithm}
\usepackage{algorithmic}
\usepackage{amsmath}
\usepackage{booktabs}
\usepackage{siunitx}
\usepackage{amssymb}
\usepackage{newfloat}
\usepackage{listings}
\DeclareCaptionStyle{ruled}{labelfont=normalfont,labelsep=colon,strut=off} 
\floatstyle{ruled}
\newfloat{listing}{tb}{lst}{}
\floatname{listing}{Listing}
\title{Enhancing Equitable Access to AI in Housing and Homelessness System of Care through Federated Learning}
    
\author{
    Musa Taib\equalcontrib, 
    Jiajun Wu\equalcontrib, 
    Steve Drew, 
    Geoffrey G. Messier
}

\affiliations {
    University of Calgary\\
    \{musa.taib, jiajun.wu1, steve.drew, gmessier\}@ucalgary.ca
}

\begin{document}

\maketitle

\begin{abstract}
The top priority of a Housing and Homelessness System of Care (HHSC) is to connect people experiencing homelessness to supportive housing. An HHSC typically consists of many agencies serving the same population.  Information technology platforms differ in type and quality between agencies, so their data are usually isolated from one agency to another. Larger agencies may have sufficient data to train and test artificial intelligence (AI) tools but smaller agencies typically do not. To address this gap, we introduce a Federated Learning (FL) approach enabling all agencies to train a predictive model collaboratively without sharing their sensitive data. We demonstrate how FL can be used within an HHSC to provide all agencies equitable access to quality AI and further assist human decision-makers in the allocation of resources within HHSC. This is achieved while preserving the privacy of the people within the data by not sharing identifying information between agencies without their consent. Our experimental results using real-world HHSC data from Calgary, Alberta, demonstrate that our FL approach offers comparable performance with the idealized scenario of training the predictive model with data fully shared and linked between agencies.
\end{abstract}

\section{Introduction}\label{sec:Intro}
Emergency housing shelters/agencies are an important part of the housing and homelessness system of care in most North American cities. While originally intended to provide low-barrier access to shelter for short periods, many people now make use of shelters over long periods either continually ({\em chronic} shelter use) or sporadically ({\em episodic} shelter use).  Most long-term shelter users face multiple physical and mental health challenges \cite{Hopper2009Shelter} that prevent them from making a quick exit from the shelter system without first connecting them with housing and the wrap-around supports necessary to help them address their unique set of challenges. Helping these people exit the shelter quickly is important. Shelters are difficult environments, and the Housing First philosophy recognizes that a person having trouble exiting a shelter should be connected to support as soon as possible \cite{goering_at_2011}.

Properly matching shelter users to supportive housing is a difficult problem. In most cities, the number of supportive housing spaces is much less \cite{ColemanDiversion2023} than the number of new people entering homelessness. The vast majority (approximately 85\%) of first-time shelter users will exit the shelter quickly and without support ({\em transitional} shelter use) \cite{aubry_identifying_2013,culhane_testing_2007}. Therefore, it is the longer-term chronic and episodic shelter users who should be prioritized for the limited supply of supportive housing. However, chronic and episodic shelter use patterns can take months or even years to become evident. As a result, most government definitions of chronic homelessness require a history of homelessness that is one year or longer \cite{byrne_testing_2015}. Requiring a person to live in a shelter for this length of time before they can demonstrate a historical record of chronic or episodic homelessness is unacceptable. Living conditions in shelters are difficult and pose further risks to a person's physical and mental health \cite{Hopper2009Shelter}.


Machine learning (ML) has great potential for rapidly identifying which first-time shelter users are at risk of becoming chronic or episodic shelter users in the long term \cite{toros_prioritizing_2018,purao_predicting_2019}. Rather than requiring a person to accumulate a record of shelter use stretching over many months or years, ML can identify at-risk people using only the first few months of a person's shelter record \cite{vanberlo_interpretable_2021}.

It must be stressed that the purpose of using ML within an HHSC is not to automate the decision of who should be assigned to supportive housing.  The established best practice within an HHSC is to support people with a variety of programs, not a single program or automated instrument \cite{shinn_allocating_2022}.  Rather than replacing front-line human staff, our tool would be used by that staff to identify possible ``under the radar'' shelter users who have been missed by other programs.  Being identified by our machine learning approach would not automatically qualify someone for housing.  Instead, it would initiate a conversation with a support worker to more fully understand that person's needs.

\begin{figure}[t]
\centerline{\includegraphics[width=3in]{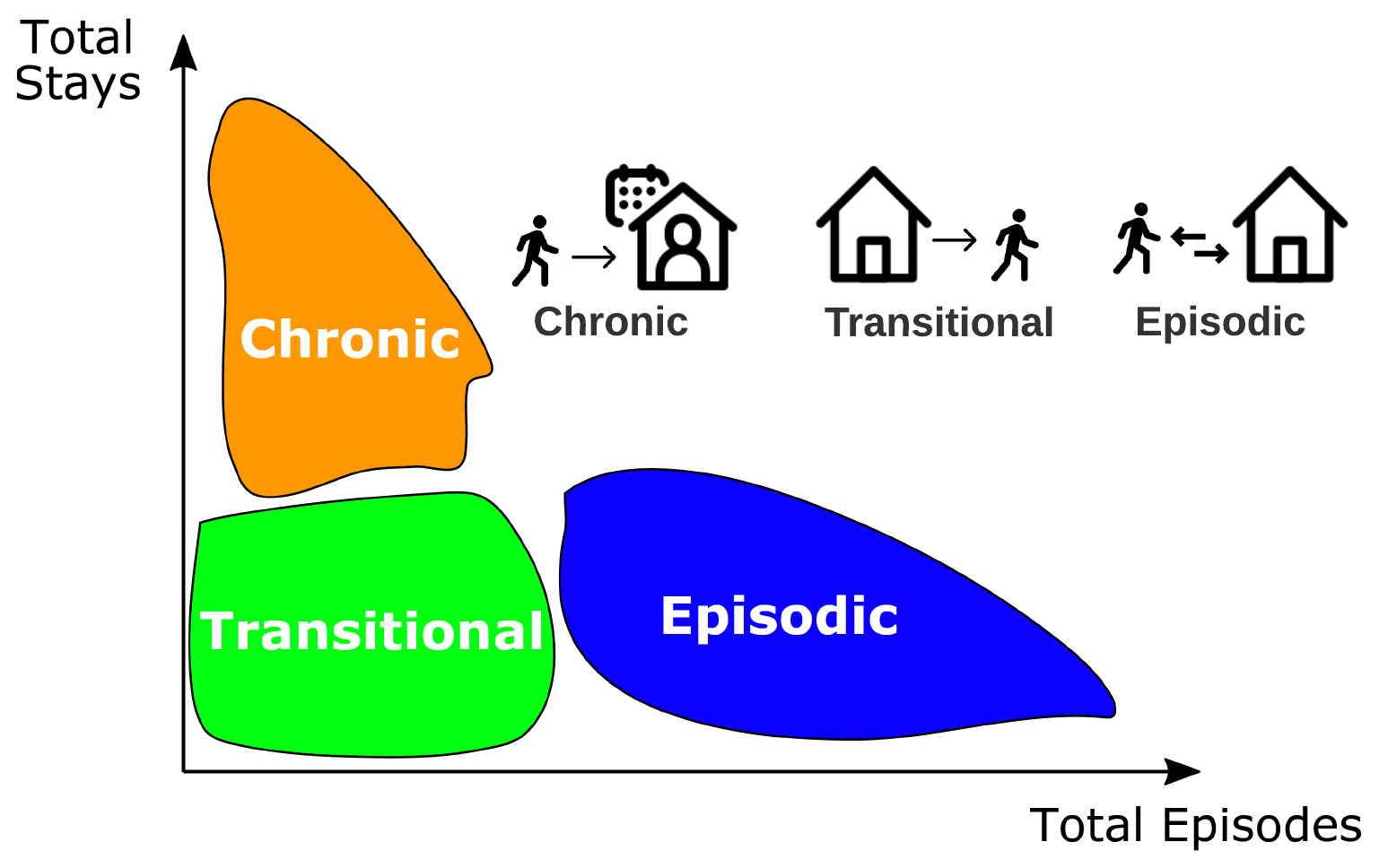}}
\caption{(Total Episode, Total Stay) clusters identifying transitional, chronic and episodic shelter use.}
\label{fg.k-means}
\end{figure}

While ML has great potential, the issue of equitable access to high-quality ML tools within an HHSC is challenging and nuanced.  An HHSC typically consists of many not-for-profit agencies of varying sizes.  Smaller agencies will usually have fewer IT resources in terms of both expertise and infrastructure.  However, IT platform and talent barriers can be overcome by the pooling of resources and/or collaboration between agency IT teams.

The more difficult problem is related to equitable access to ML due to the nature of how data exists within an HHSC. Most HHSCs are made up of a large number of government and not-for-profit community-based services. The same person may interact with multiple agencies, and each agency will record those interactions in its administrative database. Predicting the risk of episodic/chronic homelessness is most accurate when training on data that captures a person's interaction with all the agencies within an HHSC. However, maintaining a single, merged dataset for an entire HHSC is difficult.  In the spirit of being a low barrier to entry, many shelters do not require a person to sign data-sharing consent forms. Agency information technology (IT) systems are often incompatible and many municipalities also lack a single system coordinator agency with the expertise and infrastructure to merge data \cite{matthiesen2015replacing, irani2023impact}.  Finally, people are not typically assigned HHSC-wide unique identity numbers which means that merging records between agencies must be done using names and birth dates.  Sharing this identifying information across agency boundaries without consent violated the privacy of the people in the data \cite{murray_probabilistic_2015, vatsalan_taxonomy_2013, hall_privacy-preserving_2010}.  

The fragmentation of data across an HHSC means that agencies typically only have access to the data they collect by themselves.  As a result, smaller agencies that see fewer clients will have correspondingly small data sets that make it difficult to effectively train and test an ML model.  Equitable access to ML means overcoming this barrier and ensuring that all agencies have access to the same high-quality ML tools, regardless of their data set size.  At the same time, the privacy of the people in the data must be respected and identifying information must not be shared between agency boundaries without consent.

Our solution is to utilize a Federated Learning (FL) approach that collectively utilizes the disconnected datasets within the HHSC while respecting the privacy of the people in that data. Our scenario is an example of horizontal partitioning where datasets collected by each individual agency define the partitions. The model will be trained using only shelter stay data (i.e., records of when a person slept in a shelter).  While many other data features (i.e., medical history, demographics, etc.) could potentially be useful predictors, these features are not uniformly collected since not everyone utilizing an HHSC consents to providing information about themselves. There is also a risk of bias if those who do consent to providing additional information become more readily identifiable by the machine learning model. Basing our methods on patterns of shelter use only ensures that our approach can be generalized across the widest possible number of HHSCs and is as equitable as possible.

A FL perspective is also required to label the examples in the historical shelter use dataset used to train and test our model. While many government definitions exist for chronic homelessness \cite{byrne_testing_2015,messier_best_2022}, there are few established definitions for episodic homelessness. Instead, the accepted methodology for labeling a shelter use pattern as chronic, episodic, or transitional is based on the k-means cluster analysis of historical data records \cite{culhane_testing_2007,aubry_identifying_2013,kneebone_who_2015}. The k-means analysis assumes a person's entire record of homelessness is available. This requires a merged dataset of all HHSC records, which is difficult to obtain, as noted above.


The labeling process begins by using the historical shelter access records of a person to determine their number of stays and episodes. An {\em episode} is defined as a series of shelter stays with gaps of less than 30 days \cite{byrne_testing_2015}. As shown in Fig.~\ref{fg.k-means}, the k-means algorithm is instructed to create 3 clusters on a scatter plot of the two above values of everyone in the training dataset. The transitional label is assigned to people associated with the low stay/low episode cluster. The episodic labels are assigned to the low stay/high episode and chronic labels are assigned to high stay/low episode clusters, respectively. Our paper will demonstrate how to create Fig.~\ref{fg.k-means} using a federated approach that preserves privacy and does not require merging records. 

We consider three scenarios in this paper that achieve different compromises between machine learning model performance and expense:
\begin{enumerate}
\item {\bf Centralized}: All agency data has been merged by a central service. The k-means algorithm for assigning the chronic, episodic, and transitional labels to the data is run on this centralized dataset. The machine learning model is trained centrally and then deployed to the individual agencies for their use.

\item {\bf Federated}: Agencies train a single model using a federated approach that avoids the need to centrally merge data.  Labels are assigned using a federated k-means algorithm and the model is also trained using a federated approach. The resulting federated model is used by all the individual agencies.

\item {\bf Isolated}: The agencies work in isolation. Each agency runs a k-means labeling algorithm and trains its own machine learning model using only its dataset.  

\end{enumerate}

Using HHSC data from Calgary, Alberta, we demonstrate that our FL approach performs almost as well as the much less practical centralized approach. Using per-agency performance metrics, we show that smaller shelters benefit the most from our collaborative federated approach since they are unable to collect large datasets on their own. Therefore, our work addresses the important data equity issue of ensuring that all shelters have access to well-performing data tools, regardless of their size.  

Our contributions are summarized as follows:
\begin{itemize}
    \item We are the first to demonstrate how FL can be used to provide equitable access to ML tools across an HHSC with fragmented data. This is achieved without sharing data across agency boundaries or violating consent agreements. 
    \item This is the first application of FL to the k-means method of classifying shelter access patterns. While we utilize it in this paper for model training, having these labels is valuable generally for understanding how a population interacts with an HHSC.
    \item The methods in this paper are demonstrated and verified using real-world historical data from a North American HHSC.  While our specific application scenario is predicting chronic and episodic homelessness using the first few months of a person's shelter record, our framework can be generalized to any HHSC administrative data problem.  
\end{itemize}






\section{Related Work}
\label{sec:Related}
\subsection{Homelessness Prediction}
The Housing First philosophy of rapidly connecting shelter users to housing \cite{goering_at_2011,namian_governing_2020} and the prioritization of people experiencing chronic/episodic homelessness is well established \cite{aubry2015screening,gaetz_state_2016}.  While some alternative methods have been recently proposed \cite{messier_simpler_2023}, the most established method for labeling shelter access patterns as chronic, episodic or transitional is using a k-means clustering analysis of historical data records \cite{byrne_testing_2015,aubry2013,kneebone_who_2015}.  To date, much of the machine learning work related to homelessness has been concentrated on predicting a first entrance or re-entrance to homelessness \cite{clark_using_2021,diguiseppi_predictors_2020,chan_evidence_2017,kube_allocating_2019,gao_homelessness_2017,hong_applications_2018,shinn_efficient_2013,byrne_predictive_2019}. The specific prediction of chronic homelessness has also been explored \cite{vanberlo_interpretable_2021,toros_prioritizing_2018,purao_predicting_2019}.  However, all of these studies assume a single dataset.  None have explored how to overcome the practical issues that arise when working with data that has been fragmented across an HHSC.

\subsection{Federated Learning}
Federated Learning (FL) is a decentralized and privacy-preserving machine learning approach that involves training local models in each agency/device using local data, then aggregating the local models using only the model weights \cite{yang2019federated,li2020federated,lim2020federated}. 
First introduced by \citet{mcmahan2017communication}, FL enables collaborative machine learning without the need for raw data to be shared with a central authority.
In recent years, FL has seen a surge in the healthcare sector where patient data privacy is a paramount concern \cite{rieke2020future,brisimi2018federated,xu2021federated,sheller2020federated}. 
\subsubsection{FL for Risk Prediction}
Healthcare, a field highly analogous to the services provided by an HHSC, has adopted FL widely for building collaborative prediction models with administrative data \cite{brisimi2018federated,dayan2021federated}, in particular, Electronic Health Records (EHRs). For instance, FL has been used to train prediction models for cardiovascular conditions \cite{yaqoob2023hybrid}, clinical outcomes of COVID-19 patients \cite{dayan2021federated},  as well as multi-party diabetes mellitus risk prediction \cite{su2023multi}. Despite FL's capability to securely handle sensitive data, no existing studies are using FL with shelter data for risk prediction in the field of Homelessness.
This study aims to close this gap by introducing a practical FL training framework for shelters to collaboratively predict the risk of episodic and chronic homelessness without sharing the data.

\section{Dataset}
\label{sec:Data}

This study utilizes anonymized shelter data provided by the Calgary Homeless Foundation (CHF).  The protocol governing the anonymization, secure storage, and analysis of this secondary dataset was approved by the University of Calgary Conjoint Ethics Review Board (REB-19-0095).  The CHF is an organization that provides and supports a data record system utilized by most of the housing and homelessness service agencies in Calgary, Canada.  This allows the CHF to aggregate and merge records of service use for the same group of people accessing services across the system of care using privacy-preserving methods \cite{vatsalan_taxonomy_2013}.  As a result, the anonymized dataset used for this paper represents the \textit{Centralized} scenario described in the Introduction Section.  The dataset stretches from January 1, 2009 to December 31, 2019 and contains timestamped records of when a person using the HHSC slept at an emergency shelter.  There are 6,840,069 sleep records for 50,455 individuals accessing 8 different shelters.  

Restricting the machine learning model to predict episodic, chronic, and transitional shelter access patterns using sleep event data only is challenging.   There are many other data features sometimes recorded by HHSC services that may be useful in a predictive model (i.e., demographic information, medical history, employment status, family relationships, etc.).  However, it is important to stress that this information is not always collected.  Demographic information may only be recorded when accessing certain programs.  Low barrier-to-entry emergency shelters may not require clients to disclose any information to access sleep services on a cold night.
In contrast, a record of sleeping in a shelter is consistently and uniformly collected for all HHSC users.  Focusing on this data feature ensures all people are represented equitably in the data.  While specific data recording practices may vary between HHSCs, shelter sleep data is recorded by most, which means the methods presented in this paper can be easily adopted in many jurisdictions.

\subsection{Data Preprocessing}
\label{sec:Preprocessing}

The dataset utilized for this study has a series of timestamped records indicating when a person slept in an emergency shelter.  This data is pre-processed as follows:

\begin{enumerate}
    \item The sleep data records for each client are truncated after $T_o$ days.  This is the observation window used to predict whether the client will become chronic/episodic.
    \item The $T_o$ day observation window is then subdivided into $T_b$ equal segments known as time bins, where the number of sleeps in each bin is summed. For example, a client with a $T_o$ = 120-day observation window divided into $T_b$ = 10 time bins would have each bin containing the sum of their shelter stays over a period of 12 days.  Each of these time bin values is a different input data feature for the machine learning algorithm.
    \item As mentioned before our data is a series of timestamps however that does not provide much useful information to ML models. Therefore, we engineered two features to reflect the number of sleep and episodes during the observation window to increase the total number of features. We counted the number of days that a client slept at a shelter over a specified period, and an episode is defined as a period of 30 or more days between consecutive instances of sleeping.
\end{enumerate}

Note that the HHSC dataset was pre-screened by the CHF, so it was unnecessary to add preprocessing to deal with invalid data entries.  By default, the data from the CHF is linked across all HHSC agencies and corresponds to the Centralized dataset scenario.  To evaluate the Federated and Isolated scenarios, data is unlinked so that a person is assigned a new client identification (ID) number each time they access a new agency.


\subsection{Labeling}
\label{sec:Labeling}

The labeling process begins by using the historical shelter access records of an individual. A person's pattern of shelter access is labeled as transitional, chronic, or episodic by first creating the 2-tuple $(N_S,N_E)$ for that person where $N_S$ is that person's total number of shelter stays, and $N_E$ is the person's total number of shelter episodes. $N_S$ and $N_E$ are calculated using the person's first $T_p$ days of HHSC interaction.   

In a machine learning context, $T_p$ corresponds to the {\em prediction window} since this is the window over which a person's specific pattern of shelter access will manifest. In previous studies \cite{culhane_testing_2007,aubry_identifying_2013,kneebone_who_2015}, $T_p$ has been equal to the total number of days of data available for each person.  However, in the CHF data described in the Data Section, this can be several years, which is an overly ambitious prediction window given that we would like to restrict the observation window $T_o$ to only a few months. We will demonstrate in the Hyperparameter Selection Section that reducing $T_p$ to be on the order of 1-2 years improves performance while still providing a very impressive prediction horizon.

Once the 2-tuples are determined for each person in the data, the k-means algorithm with 3 clusters is used to associate each person with a centroid that corresponds to transitional, chronic, and episodic shelter use as shown in Fig.~\ref{fg.k-means}.  For the Centralized scenario, the 2-tuples are created for everyone in the data using the fully merged CHF dataset.  For the Isolated case, a person's interaction with multiple agencies is unlinked and treated as different individuals as described in the Data Preprocessing Section.  Each agency performs its own k-means labeling process for the unlinked individuals in their own datasets.

For the Federated scenario, we propose a new labeling approach, Decentralized k-means, as shown in Algorithm \ref{alg:FLlabel}, that allows agencies to benefit from pooling their data without the technical complexity and privacy concerns of fully linking their datasets. Each agency performs k-means clustering on its own datasets for the Isolated case. The clusters are then forwarded to a central location that performs a weighted average. These averaged clusters are then shared back with the agencies who can use them to re-classify their 2-tuples.

\begin{algorithm}[t]
    \caption{\texttt{DecentralizedLabeling}}
    \label{alg:FLlabel}
    \textbf{Input}: $\mathcal{K}$ is the list of agencies; $k \in \mathcal{K}$ is an agency; $|\mathcal{D}_k|$ is the size of the dataset for agency $k$. $\mu_c^k, \mu_e^k, \mu_t^k$ are chronic, episodic, and transitional cluster centroids on agency $k$; $\mu_c, \mu_e, \mu_t$ are global chronic, episodic, and transitional cluster centroids used for labeling.   \\ 
    \textbf{Output}: $\mu_c, \mu_e, \mu_t$. 

\begin{algorithmic}[1]
\FOR {each agency $k \in \mathcal{K}$ } 
    \STATE Initialize $\mu_c^k, \mu_e^k, \mu_t^k$ from $\mathcal{D}_k$    
    \STATE $\text{Apply k-means clustering for }\mathcal{D}_k$
    \STATE Update $\mu_c^k, \mu_e^k, \mu_t^k$ 
    \ENDFOR
\STATE $\mu_c, \mu_e, \mu_t \gets 
\sum_{k=1}^{\mathcal{K}}
\frac{\lvert \mathcal{D}_k \rvert}{\sum_{i=1}^\mathcal{K} \lvert \mathcal{D}_i \rvert}
\mu_c^k, \mu_e^k, \mu_t^k$

\STATE \textbf{return} $\mu_c, \mu_e, \mu_t$

\end{algorithmic}   

\end{algorithm}

\section{Methodology}
\label{sec:Methods}
\subsection{Problem Formulation}



We formulate the allocation of the homelessness support problem as a multi-class prediction task. Consider a dataset $\mathcal{D}$, which is a $m \times n$ matrix where $m$ is the number of clients and $n$ is the number of features. The sleep pattern of an individual client is split into $T_b$ time bins in addition to two engineered features: total number of sleeps and total number of episodes.
Therefore, each client would have an input feature vector $\mathbf{f}$ with length $n$ = $T_b+2$. Our task is to train a classifier $C(.)$, which produces the prediction results for a given $\mathbf{f}$; $\hat{y} = C(\mathbf{f})$, where $\hat{y} \in $ [chronic, episodic, transitional] refers to the output labels. 




\begin{algorithm}[t]
    \caption{\texttt{AgencyTraining}. The Local Prediction Model Training for each agency.}
    \label{alg:algorithm1}
    \textbf{Input}: $w$ is the machine learning model; $\eta$ is the learning rate; $\mathcal{E}$ is the number of training epochs; $B$ is the training batch size; $\mathcal{L}$ is the loss function of training.   \\
    \textbf{Output}: The trained agency model.

\begin{algorithmic}[1]

\STATE $\mathcal{B} \leftarrow \text{split }\mathcal{D}_k\text{ into batches of size }B$
    \FOR{each local epoch $i$ from $1$ to $\mathcal{E}$}
        \FOR {each batch ${ b \in \mathcal{B}}$}
        \STATE $w \gets w - \eta \nabla
        \mathcal{L} (w;{b})$
        \ENDFOR
    \ENDFOR
\STATE \textbf{return} $w$
    
\end{algorithmic}   

\end{algorithm}

\begin{figure*}[t]
    \centering
    \includegraphics[width=\textwidth]{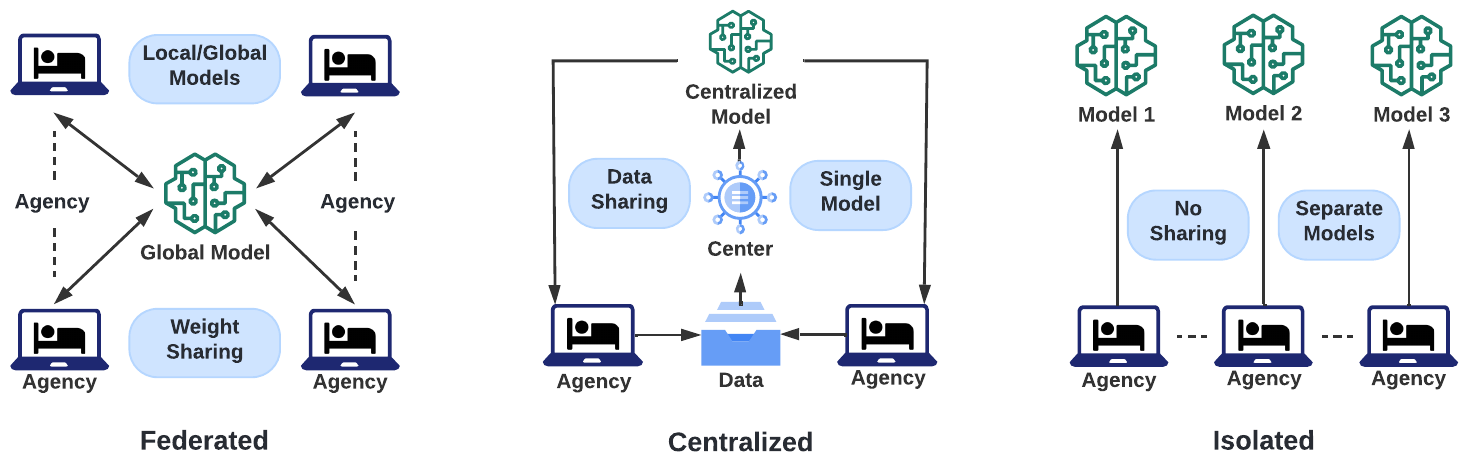}
    \caption{Illustration of the three scenarios for training predictive models in the paper: Federated, Centralized, and Isolated.}
    \label{fig:3in1framwork}
\end{figure*}

\subsection{Training Details}

Fig.~\ref{fig:3in1framwork} shows the training scenarios of the three models. Training for Centralized and Isolated scenarios involves the preprocessing steps described in the Data Preprocessing Section, dataset labeling as described in the Labeling Section, and updating the machine learning model as shown in Algorithm~\ref{alg:algorithm1}. The only difference is that Centralized and Isolated scenarios utilize linked and unlinked data described in the Data Preprocessing Section, respectively.

Our proposed FL framework is based on the FedAvg algorithm, following the steps below:
\begin{itemize}
    \item \textbf{Step 1}: Apply Decentralized k-means from Algorithm~\ref{alg:FLlabel} to generate the labels for the training data.
    \item \textbf{Step 2}: The central server broadcasts the global model to $\mathcal{K}$ available local agencies.
    \item \textbf{Step 3}: Each agency updates and uploads its local model to the server after local training.
    \item \textbf{Step 4}: The server aggregates all uploaded local models into the new global model for the next round. Repeat \textbf{Steps 2-4} until the global model converges.
    
\end{itemize}

The pseudocode for the FL scenario is shown in Algorithm~\ref{alg:algorithm2}.

\begin{algorithm}[t]
    \caption{Federated Prediction Model Training}
    \label{alg:algorithm2}
    \textbf{Input}: $\mathcal{K}$ is the list of agencies; $\mathcal{D}$ is the list of dataset for all agencies; $|\mathcal{D}_k|$ is the size of the dataset for agency $k$; $\mathcal{T}$ is the number of communication rounds; $w_{t}^k$ is the global model for agency $k$ at round $t$; $\mathcal{B}$ is the batch size; $\theta$ is the clusters.   \\
    \textbf{Output}: The trained global model parameters. 

\begin{algorithmic}[1]

\STATE $\texttt{DecentralizedLabeling}(\mathcal{K}, \mathcal{D}$) 
\hspace{11mm} 
$\triangleright$ 
\textbf{Alg. \ref{alg:FLlabel}} 

\STATE Server initialize $w_0$;

\STATE \textbf{Training Starts: }
\FOR {each round $t$ from $1$ to $\mathcal{T}$}

    \FOR{each agency $k \in \mathcal{K}$ in parallel}
        \STATE $w_{t}^k 
        \gets
        \texttt{AgencyTraining}(k, \mathcal{D}_k, w_{t-1})$  
        $\triangleright$  
        \textbf{Alg.  \ref{alg:algorithm1}}
    \ENDFOR
    \STATE $w_{t+1} \gets  
    \sum_{k=1}^{\mathcal{K}}
    \frac{\lvert \mathcal{D}_k \rvert}{\sum_{i=1}^{\mathcal{K}} \lvert \mathcal{D}_i \rvert}
    w_{t}^k $
\ENDFOR 

 
\end{algorithmic}   

\end{algorithm}

\section{Evaluation}
\label{sec:Evaluation}
We explain the experimental setup of our 3 proposed methods: Centralized, Isolated, and Federated.  

\subsection{Experimental Setup}
The experiments were conducted in a simulated setup on an Apple M1 Pro MacBook with 16GB of RAM \footnote{The details of the model and code are available at https://github.com/MusaJTaib/FederatedLearning}. The neural networks trained at each simulated agency were isolated to ensure privacy for isolated and federated models. There was no direct inter-agency communication except for controlled weight sharing between training rounds in federated cases. There was no communication for isolated cases.



\subsection{Train/Test Splits}


We apply a normalization process to the dataset using z-scoring \cite{scikit-learn}. 
This results in a standardized score for each data point, reflecting how many standard deviations it is from the column average. 

Following the normalization, the dataset is divided into stratified training and testing sets, adhering to an 80/20 split. This strategy guarantees that both sets maintain proportional representations of clients across different agencies and labels. For the Federated and Isolated models, the test set labels are synchronized with those from the centralized model.

\subsection{Machine Learning Models}
Multi-layer Perceptrons (MLPs) are a class of neural networks characterized by their stacked arrangement of fully connected layers, which makes them particularly adept at extracting and learning complex patterns from the input data \cite{Goodfellow-et-al-2016}. The MLP model used for this paper features three ReLU-activated hidden layers \cite{Glorot2011} of sizes 512, 128, and 16, forming a structure capable of identifying and processing complex patterns within the dataset. This configuration, with its hierarchical design, is adept at capturing and learning the intricate relationships present in the input data. To facilitate model generalization and curtail overfitting, dropout layers with rates ranging from 0.1 to 0.4 are judiciously applied across the dense layers \cite{Srivastava2014}. In the final output layer, a sigmoid function \cite{Goodfellow-et-al-2016} is employed to handle multi-class classification.

The training for each model is performed as described in Algorithm \ref{alg:algorithm1}. The models employed the Adam optimizer \cite{Kingma2014} with a learning rate $\eta$ = 0.02. For loss calculation, categorical cross-entropy is selected, aligning with the multi-class nature of the output. Optimization is performed via mini-batch gradient descent with a batch size $\mathcal{B}$ = 500 and the models are trained for a total of 200 epochs ($\mathcal{E}$).

The performance of the model can vary as a result of random weight initializations. Therefore, each MLP model was trained 10 times on different initializations for all three cases (federated, isolated, and centralized). The results for all 10 rounds were then averaged to get a more generalized picture of the results.

The aforementioned details apply uniformly across all three models, with the federated model incorporating additional parameters to tailor the training process. According to Algorithm \ref{alg:algorithm2}, the federated model undergoes $\mathcal{T}$ = 75 communication rounds. With $\mathcal{K}$ = 8 each representing a unique agency contributing to the model. Furthermore, each client undergoes $\mathcal{E}$ = 15 training epochs on their local dataset for every round.

\begin{table}[htb]
\centering
\begin{tabular}{ccccc}
\toprule
$T_b$ (Bins) & $T_o$ (Days) & $T_p$ (Days) & Precision & Recall \\
\midrule
 & & 120 & 0.618 & 0.735 \\
 & 548 & 90 & 0.568 & 0.691 \\
 & & 60 & 0.529 & 0.648 \\
\cmidrule{2-5}
 & & 120 & 0.584 & 0.696 \\
 5 & 730 & 90 & 0.547 & 0.654 \\
 & & 60 & 0.504 & 0.609 \\
\cmidrule{2-5}
 & & 120 & 0.549 & 0.666 \\
 & 913 & 90 & 0.514 & 0.631 \\
 & & 60 & 0.479 & 0.593 \\
\midrule
 & & 120 & 0.622 & 0.729 \\
 & 548 & 90 & 0.571 & 0.683 \\
 & & 60 & 0.530 & 0.634 \\
\cmidrule{2-5}
 & & 120 & 0.608 & 0.686 \\
10 & 730 & 90 & 0.545 & 0.656 \\
 & & 60 & 0.499 & 0.604 \\
\cmidrule{2-5}
 & & 120 & 0.560 & 0.658 \\
 & 913 & 90 & 0.511 & 0.632 \\
 & & 60 & 0.481 & 0.593 \\
\bottomrule
\end{tabular}
\caption{Precision and Recall values for Centralized model with different time bins ($T_b$), observations ($T_o$) and prediction ($T_p$) windows.}
\label{tab:centralized_model}
\end{table}






\subsection{Hyperparameter Selection} 
\label{sec:Hyperparameters}

The Data Preprocessing and Labeling Sections present some important hyperparameters that have a significant impact on model performance.  Increasing the size of the observation window, $T_o$, will generally improve model performance.  However, a long observation window is at odds with our objective of identifying at-risk HHSC users as soon as possible.  Therefore, the smallest possible $T_o$ that yields acceptable performance must be selected.

Increasing the number of time bins, $T_b$, improves the model's ability to capture temporal trends and provide a better fit to the data.  In general, a small value of $T_b$ will underfit and a large value of $T_b$ will overfit the data \cite{taib_efficient_2023}.  

Finally, model performance will improve with a decreasing prediction window $T_p$ since a smaller $T_p$ improves the relationship between the labels and the pattern of shelter use early in a person's HHSC record.  However, $T_p$ should still be made as large as possible to allows the longer term patterns of chronic and episodic homelessness to develop.

Table~\ref{tab:centralized_model} shows model performance for the Centralized scenario for various values of $T_o$, $T_b$ and $T_p$.  As noted above, the best performance is achieved with the largest values of $T_o$ and $T_b$ and the smallest value of $T_p$.  In the remainder of the paper, the values of $T_b$ = 10 and $T_p$ = 548 days will be used.  However, a value of $T_o$ = 90 days will be used instead of 120 days to achieve a compromise between performance and rapid identification.

\subsection{Comparing Centralized, Federated and Isolated Prediction}
\label{sec:ModelComparison}

Using the hyper-parameters from the Hyperparameter Selection Section, Table~\ref{tab:performance_evaluation} shows precision and recall for all three data scenarios averaged across 10 randomized initializations.

\begin{table}[t]
    \centering
    \begin{tabular}{lcccc}
        \toprule
        Model       & Linked Data & Precision & Recall \\
        \midrule
        Centralized & \checkmark         & 0.5713    & 0.6826 \\
        Federated   &                    & 0.5379    & 0.6495 \\
        Isolated    &                    & 0.5061    & 0.4656 \\
        \bottomrule
    \end{tabular}
    \caption{Prediction performance for the Centralized, Federated, and Isolated scenarios.}
    \label{tab:performance_evaluation}
\end{table}

\begin{table}[t]
    \centering
    \begin{tabular}{lccc}
        \toprule
        Model   & Transitional  & Episodic   & Chronic\\
        \midrule
        C-P     & 0.90          & 0.43       & 0.30   \\
        F-P     & 0.92          & 0.27       & 0.41   \\
        I-P     & 0.80          & 0.28       & 0.42   \\
        C-R     & 0.82          & 0.43       & 0.72   \\
        F-R     & 0.68          & 0.59       & 0.69   \\
        I-R     & 0.74          & 0.45       & 0.20   \\
        \bottomrule
    \end{tabular}
    \caption{Prediction performance per class for the Centralized, Federated, and Isolated scenarios. C-, F-, and I- denote Centralized, Federated, and Isolated respectively. P: precision, R: recall.}
    \label{tab:performance_evaluation_perclass}
\end{table}

As expected, the Centralized scenario achieves the best performance in Table~\ref{tab:performance_evaluation} due to the model being able to utilize a fully linked dataset.  However, the Federated model achieves performance that comes within 4\% of the Centralized model.  This is excellent performance, especially since the federated methods avoid the privacy and technical barriers of linking data.  

Looking at Table~\ref{tab:performance_evaluation_perclass} we can also see another advantage of using FL over the current local shelter implementation. The per class precision and recall for the FL case are in general much better than for local classes and fall in line with previous trends. This is quite important for this case because of the highly imbalanced nature of the dataset. As overall higher results shown before could have simply been attributed to a single class’s performance improvement. However, it is quite obvious that clients in all three classes are seeing the benefits of this improvement making the model even more fairer to use compared to the current local implementations.

Performance achieved by the Isolated scenario is far below Federated and Centralized, especially in terms of recall.  This shortfall can be explained by the per-agency Federated and Isolated scenario performance results presented in Table~\ref{tab:FLVsISO}.  Each row in this table includes the agency ID number, the number of individuals in that agency's dataset, and the precision and recall achieved by the Federated and Isolated scenarios.

Table~\ref{tab:FLVsISO} demonstrates that it is the smaller shelters (shelters with capacity below 12,000) that perform particularly poorly in the Isolated case due to the very small size of their datasets. An exception to this trend is noted in the case of agency 213. Given its limited size (293 clients) and the consequently small test sample (approximately 70 clients), the results for this agency are deemed insufficient for a reliable evaluation of model performance and are considered an outlier. 

\begin{table}[ht]
\centering
\sisetup{table-format=1.3} 
\begin{tabular}{
    S[table-format=3.0] 
    S[table-format=5.0] 
    S 
    S 
    S 
    S 
}
\toprule
{Agency} & {Clients} &{ F-P }& {F-R }& {I-P} & {I-R} \\
\midrule
4   & 2386  & 0.411 & 0.641 & 0.203 & 0.397 \\
13  & 7063  & 0.443 & 0.659 & 0.343 & 0.388 \\
55  & 12017 & 0.502 & 0.67  & 0.57  & 0.411 \\
188 & 15065 & 0.605 & 0.549 & 0.499 & 0.452 \\
213 & 293   & 0.381 & 0.508 & 0.38  & 0.636 \\
225 & 1207  & 0.515 & 0.536 & 0.249 & 0.386 \\
330 & 28523 & 0.581 & 0.716 & 0.629 & 0.568 \\
333 & 11713 & 0.473 & 0.607 & 0.341 & 0.353 \\
\bottomrule
\end{tabular}
\caption{Experiments on the real-world dataset. F- and I- denote Federated and Isolated, respectively. P: precision, R: recall.}
\label{tab:FLVsISO}
\end{table}

\begin{figure}[t]
    \centering
    \includegraphics[width=0.99\linewidth]{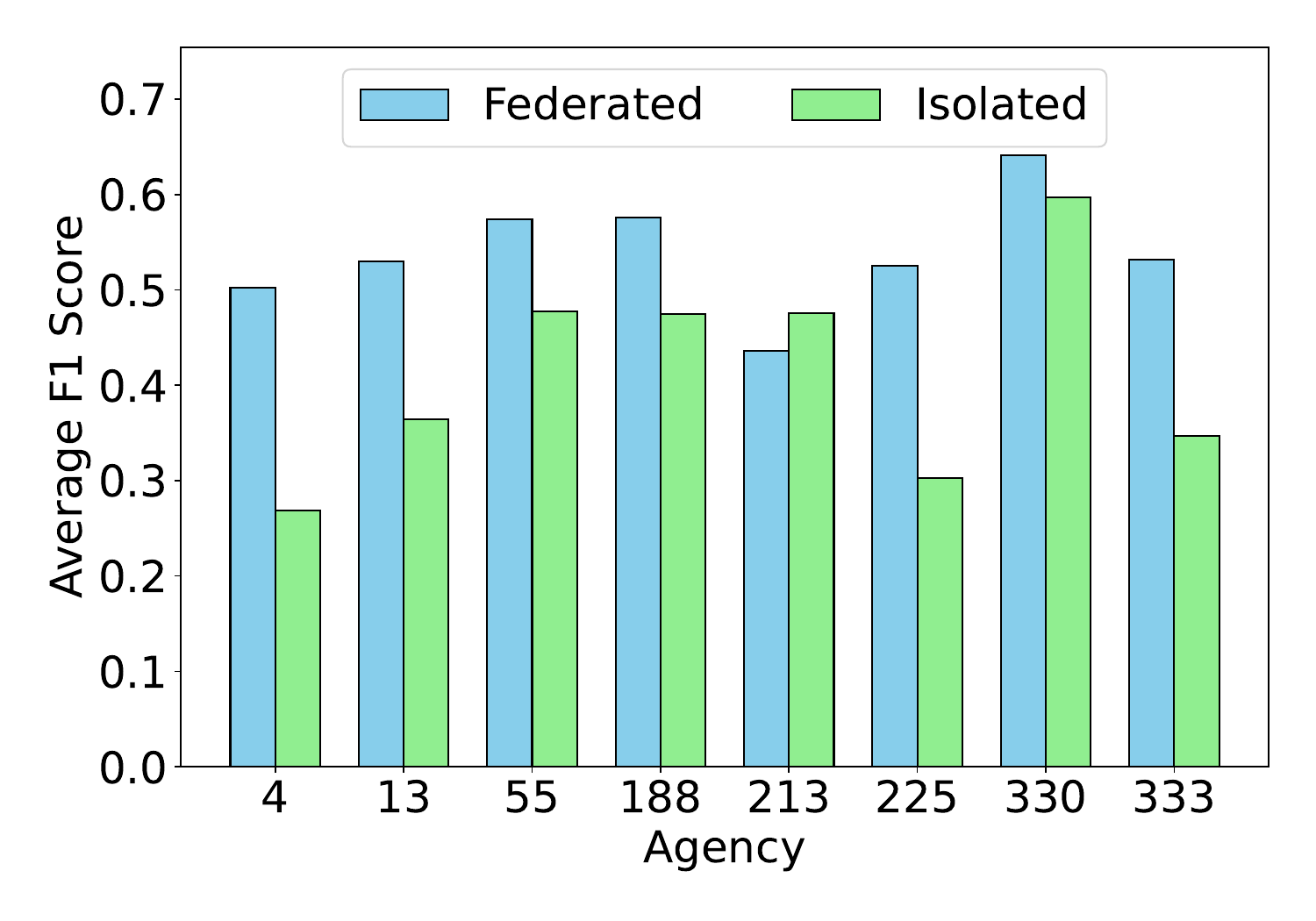}
    \caption{F1 Scores for Federated and Isolated cases} \label{fig:F1FVI}
\end{figure}

Figure \ref{fig:F1FVI} compares the Federated and Isolated cases using the F1 score.  In this figure, it is evident that agencies with large datasets (notably 330, 188, and 55) can achieve respectable performance using their own datasets in isolation.  This further underscores that the Federated approach has the largest impact on shelters with smaller operations.  



\section{Discussion}
\label{sec:Discussion}

Before discussing the results in the Evaluation Section, it is important to provide context regarding how a machine learning tool should be utilized within an HHSC.  As noted in the Introduction Section, machine learning should never be the only method for connecting people to housing.  Instead, it is used as part of a range of programs for engaging with HHSC users \cite{shinn_allocating_2022}, including ``diversion'' programs specifically designed for making contact with new HHSC entrants.

While improving the absolute performance of the models presented in the Evaluation Section is always desirable, these performance levels are acceptable if the models are used in the context of supporting a robust diversion program with human front-line staff. Performance could also be further improved if the methods in this paper are applied in an HHSC where other data features, such as demographic information, are collected.

\subsubsection{Centralized vs. Federated}

The results from Table \ref{tab:performance_evaluation} show that the centralized model outperforms the other two models.  The Centralized scenario is ideal and may be achievable in some HHSCs that have a centralized agency capable of collecting data, as is the case in Calgary.  However, most HHSCs are better represented by the Isolated scenario where agencies would need to operate independently without collaboration. The Federated approach, as proposed in this paper, is posited as a viable compromise, offering a balance between prediction performance and minimizing the amount of centralized IT infrastructure required.  It can also be implemented without requiring the people using the HHSC to provide consent to share data across agency boundaries.

\subsubsection{Achieving Equity with Federated Learning}

One significant challenge within the HHSC network is the disparity in data analytics capabilities between large and small agencies. Smaller agencies often struggle with insufficient datasets. This limitation hampers their ability to identify at-risk clients accurately and tailor their services effectively. Looking at the results in Figure \ref{fig:F1FVI}, it is quite clear that our federated model has a substantial gain in performance over the isolated case. A breakdown of the per agency performance keeping in mind the number of clients per agency in table \ref{tab:FLVsISO} shows exactly where the federated model gains its performance. Notably, smaller shelters, those with fewer than 10,000 clients, which struggle to adequately train their models due to insufficient data, benefit markedly from the aggregated insights available through larger shelters. Conversely, larger shelters observed minimal performance gains, as their pre-existing data volumes were already sufficient to effectively train their models. 

The federated model presents a solution to this problem by enabling smaller agencies to leverage larger, aggregated models trained across multiple data sources without directly sharing sensitive information. FL allows for the collective improvement of model performance, benefiting all participants in the network. When using machine learning for social good, equity is important. This means that all people accessing an HHSC should benefit from the same high-quality tools regardless of whether they choose to interact with a large or small agency.  The fact that our proposed FL approach benefits smaller agencies, in particular, shows it is an important tool for achieving this equity goal.

\subsubsection{Federated models reduce the need for data linkage}
The conventional approach to enhancing machine learning models across different agencies involves the linkage of data to create a centralized dataset. This process not only raises significant privacy concerns but also involves logistical complexities and substantial resources dedicated to ensuring the secure and accurate merging of sensitive information. The federated model offers a compelling advantage by eliminating the necessity for direct data linkage. By training models on decentralized data and aggregating the learned models, FL circumvents the challenges associated with data consolidation. This approach significantly reduces the complexity and cost associated with data linkage, reallocating those resources toward improving service delivery and operational efficiencies. Moreover, FL upholds the privacy and confidentiality of individual records, addressing one of the critical barriers to data sharing among agencies.


\section{Conclusion}
\label{sec:Conclusion}

In this work, a critical gap in current homelessness management strategies is addressed by using a machine learning approach that respects the privacy and autonomy of individual agencies while leveraging data from across the HHSC. In jurisdictions where raw data sharing across agencies is challenged by privacy concerns or logistical challenges, our proposed FL model demonstrated virtually the same performance as that of a centralized data approach. Compared to the status quo in the domain, our federated approach greatly enhances the ability of smaller shelters to make predictions and also promotes equity across the HHSC. By demonstrating the feasibility and effectiveness of federated learning in this context, this study lays the groundwork for future research and development of ML applications within the social services sector. 

\section{Acknowledgements}
The authors would like to acknowledge the support of Making the Shift, the Calgary Homeless Foundation, and the Government of Alberta. This study is based in part on data provided by Alberta Seniors, Community and Social Services. The interpretation and conclusions contained herein are those of the researchers and do not necessarily represent the views of the Government of Alberta. Neither the Government of Alberta nor Alberta Seniors, Community, and Social Services express any opinion related to this study. This research is supported in part by the Natural Sciences and Engineering Research Council of Canada (NSERC) Discovery Grant RGPIN-2024-03954.

\bibliography{Administrative_Data, Musa_References, ML_Fundamentals, Sequences_and_Sessions, Health, Signal_Processing, Homelessness_AI, Homelessness_Patterns, messier_zotero, Gerry-Ref}

\end{document}